\documentclass{l4dc2027}
\title[SOC-Net]{
SOC-Net: Second-Order-Cone-Constrained Neural Networks 
}

\usepackage{times}
\usepackage{algorithm}
\usepackage{algpseudocode}
\usepackage{array,booktabs}
\usepackage{graphicx}
\usepackage{placeins}

\newtheorem{assumption}{Assumption}

\newcommand{\nin}{n_{\mathrm{in}}}
\newcommand{\nout}{n_{\mathrm{out}}}
\newcommand{\R}{\mathbb{R}}
\newcommand{\Psoc}{\mathcal P_{\mathrm{SOC}}}

\author{%
    \Name{Seungseok Oh\nametag{\thanks{These authors contributed equally.}}} \Email{seungseok.oh@snu.ac.kr}\\
    \addr Seoul National University, Seoul, South Korea\\
    \Name{Youngjae Min\nametag{\footnotemark[1]}} \Email{yjm@mit.edu}\\
    \addr Massachusetts Institute of Technology, Cambridge, MA, USA\\
    \Name{Navid Azizan} \Email{azizan@mit.edu}\\
    \addr Massachusetts Institute of Technology, Cambridge, MA, USA\\ %02139, United States
    \Name{Namhoon Cho}\thanks{Corresponding author} \Email{namhoon.cho@snu.ac.kr}\\
    \addr Seoul National University, Seoul, South Korea
}

\begin{document}

\maketitle
\thispagestyle{plain}

\vspace{-3em}

\begin{abstract}
    In many applications, neural networks must learn input--output maps that respect prior knowledge such as physical laws, safety requirements, and operating limits.
    These requirements constrain which outputs are admissible for a given input and often admit second-order cone (SOC) representations, including norm bounds and robust linear inequalities. Examples in control include friction cones, thrust bounds, and input-robust control barrier function (CBF) conditions.
    % Recent work builds neural networks whose outputs satisfy hard constraints by construction, even outside the training distribution.
    However, existing approaches to enforcing general input-dependent SOC constraints require a conic solver, an iterative scheme whose feasibility depends on convergence, or an interior feasible point for each input.
    We propose SOC-Net, a neural network architecture that enforces input-dependent SOC constraints by construction for every input, even outside the training distribution.
    SOC-Net projects an affine lifting of the backbone output onto the reachable slice of a fixed product of Lorentz cones and maps the correction back to the output space. For separable constraints, this projection splits into one projection per cone, and each is available in closed form or reduces to a single scalar equation, regardless of the cone dimension.
    We prove that SOC-Net retains the expressive power of the backbone network under regularity conditions and accommodates end-to-end training by standard backpropagation with an explicit Jacobian.
    Experiments on contact force allocation with friction constraints and 3-D obstacle avoidance with an input-robust CBF demonstrate that SOC-Net satisfies the constraints at every evaluated input, with lower evaluation latency than the tested solver-based and iterative baselines.
\end{abstract}

\begin{keywords}
    Hard-constrained neural networks, HardNets, second-order cone constraints, learning to optimize, learning-based control, constrained control
\end{keywords}

\section{Introduction}
\label{sec:introduction}

As neural networks proliferate across scientific and engineering applications, they are increasingly required to learn input--output maps that respect prior knowledge such as physical laws, safety requirements, and operating limits.
This knowledge restricts which outputs are admissible for a given input.
In control, such networks now serve as dynamics models \citep{shi2019neural}, feedback policies \citep{levine2016end}, and optimization-based controllers \citep{chen2018approximating} of physical systems.
Because their outputs affect a physical system, the knowledge must also hold at deployment, where a single inadmissible output can be unsafe.

Many of these requirements and limits can be written as second-order cone (SOC) constraints $\|Ay+b\|_2\le c^\top y+d$ on the output $y$.
SOC constraints express norm bounds and robust linear inequalities \citep{lobo1998applications}.
Friction cones in contact force allocation \citep{boyd2007fast} and thrust and pointing limits in powered-descent guidance \citep{acikmese2007convex} are SOC constraints.
So are robust control barrier function (CBF) conditions, whose safety margin grows with the model uncertainty \citep{dean2021guaranteeing,cosner2021measurement,taylor2021towards}.
In learning problems the coefficients $A,b,c,d$ usually depend on the input $x$, so the feasible set changes from one evaluation to the next.

A common way to impose such constraints is to penalize their violation in the training loss.
Examples include fixed penalties, physics-informed residuals, augmented-Lagrangian terms, and primal--dual constrained learning \citep{raissi2019physics,dener2020training,chamon2020probably,fioretto2020predicting}.
These methods reduce violations on the training distribution but do not guarantee that every output is feasible.
Violations can remain on unseen inputs or before training has converged.
%DC3 \citep{donti2021dc3} adds gradient-based correction steps to the output, but a finite number of steps does not guarantee inequality feasibility either.

Hard-constrained neural networks (HardNets) instead guarantee feasibility for every input by construction. %by projecting the backbone's output to a feasible point.
HardNet-Cvx \citep{min2024hardnet} enforces input-dependent convex constraints by projecting network outputs onto the feasible set using differentiable optimization layers \citep{amos2017optnet,agrawal2019differentiable}. %achieve this by solving a constrained problem in every forward pass. Hard-constrained neural networks avoid the solver by mapping the backbone output to a feasible point explicitly.
Other hard-constrained architectures avoid a general-purpose solver by mapping the backbone output to a feasible point explicitly.
For instance, HardNet-Aff \citep{min2024hardnet} does so in closed form through metric projection but handles only affine constraints.
ECO \citep{goertzen2026eco} develops a closed-form projection layer for convex quadratic constraints.
RAYEN supports SOC and LMI constraints through rays from an interior feasible point \citep{tordesillas2023rayen}.
When the constraint depends on the input, this point must be supplied for every input.
To avoid that, LMI-Net and $\Pi$Net rely on iterative splitting or projection \citep{tang2026lmi,grontas2026pinet}.
Their outputs are feasible only upon convergence, so truncated iterations may return infeasible outputs.
HardNet++ \citep{goertzen2026hardnet++} addresses nonlinear equality and inequality constraints through differentiable, damped local linearizations.
For general input-dependent SOC constraints, existing methods therefore rely on a conic solver, an iterative scheme, or an interior point computed for each input.

\begin{figure}[t]
    \centering
    \includegraphics[width=1.0\linewidth]{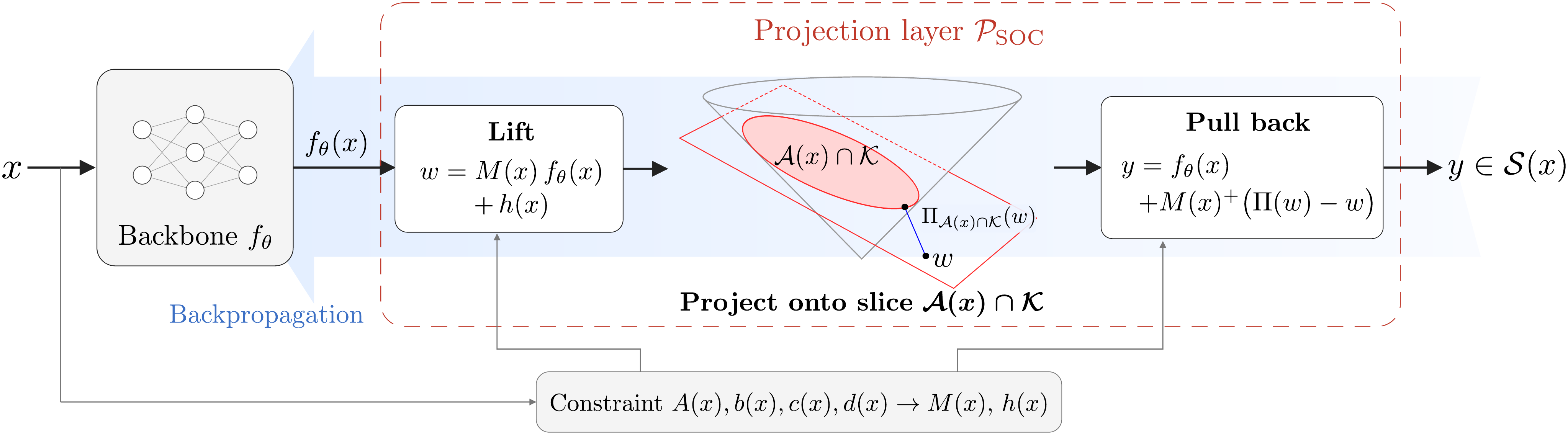}
    \caption{Schematic of SOC-Net for a single-constraint case. The backbone output $f_\theta(x)$ is lifted to $w=M(x)f_\theta(x)+h(x)$, projected within the reachable affine subspace $\mathcal A(x)$ onto the slice $\mathcal A(x)\cap\mathcal K$, and pulled back through $M(x)^+$, so that $y\in\mathcal S(x)$ for every input $x$.}
    \label{fig:pipeline}
\end{figure}

We propose SOC-Net, a neural network architecture that enforces hard, input-dependent SOC constraints on its output.
SOC-Net rewrites the constraints as $M(x)y+h(x)\in\mathcal K$ for a fixed product $\mathcal K$ of Lorentz cones and projects the lifted backbone output onto the part of $\mathcal K$ reachable by some output (Fig.~\ref{fig:pipeline}).
When the constraints are separable, as for friction cones at distinct contacts, this projection splits into one projection per cone.
When $M(x)$ is surjective, this is the classical Lorentz-cone projection.
When it is not, we show that projecting onto the reachable slice reduces to two coordinates and one monotone multiplier equation.
This reduction holds regardless of the cone dimension and the rank of $M(x)$.
The correction therefore needs neither a conic solver nor a feasible anchor.
It also remains feasible for any approximate root, unlike splitting layers whose feasibility is tied to convergence.

Our contributions are as follows.
\begin{itemize}
    \item We propose SOC-Net, a neural network architecture whose output satisfies input-dependent SOC constraints by construction for every input. It needs only a closed-form projection or a single scalar root per constraint.
    \item We prove that SOC-Net is feasible for every input even with an inexact root and leaves feasible outputs unchanged. Under explicit regularity conditions on the moving slice, it also retains universal approximation over continuous feasible targets. Its explicit Jacobian requires no iterations, so the whole network can be trained end-to-end by standard backpropagation.
    \item We evaluate SOC-Net on contact force allocation and on robust CBF-constrained control. Baselines include soft-penalty networks, other hard-constrained layers, and conic solvers.
\end{itemize}

\section{Problem Formulation}
\label{prob_form}

Consider a neural network $f_\theta:\mathcal X\subseteq\R^{\nin}\to\R^{\nout}$, parameterized by $\theta$.
We seek a constrained network $x\mapsto y\in\R^{\nout}$, built from $f_\theta$, whose output satisfies the input-dependent SOC constraints
\begin{equation}
\|A_i(x)y+b_i(x)\|_2\le c_i(x)^\top y+d_i(x),\qquad i=1,\ldots,p, \quad \forall x \in \mathcal X,
\label{eq:soc}
\end{equation}
where $A_i(x)\in\R^{k_i\times \nout}$, $b_i(x)\in\R^{k_i}$, $c_i(x)\in\R^{\nout}$, and $d_i(x)\in\R$; let $\mathcal S(x)\subseteq\R^{\nout}$ denote the feasible output set of \eqref{eq:soc} at $x$.

For the $i$th constraint, define the lifting matrix and offset
\begin{equation}
    \label{eq:soc-lifting}
    M_i(x)=\begin{bmatrix}c_i(x)^\top\\A_i(x)\end{bmatrix}\in\R^{(k_i+1)\times\nout},
    \qquad
    h_i(x)=\begin{bmatrix}d_i(x)\\b_i(x)\end{bmatrix}\in\R^{k_i+1},
\end{equation}
and the Lorentz cone $\mathcal K_i=\{(t,z)\in\R\times\R^{k_i}:\|z\|_2\le t\}$, whose apex is the origin and whose axis is $e_i=(1,0,\ldots,0)^\top$.
Then an output $y$ satisfies the $i$th constraint of \eqref{eq:soc} if and only if its lifted image $M_i(x)y+h_i(x)$ lies in $\mathcal K_i$.
We stack the liftings vertically as
\begin{equation}
    \label{eq:multi-soc-stack}
    M(x)=\bigl[M_1(x);\ldots;M_p(x)\bigr],
    \qquad
    h(x)=\bigl[h_1(x);\ldots;h_p(x)\bigr],
\end{equation}
where $[\,\cdot\,;\,\cdot\,]$ denotes vertical concatenation.
Then \eqref{eq:soc} holds if and only if $M(x)y+h(x)$ lies in the product cone $\mathcal K=\mathcal K_1\times\cdots\times\mathcal K_p$.
We call $\mathcal A(x)=h(x)+\operatorname{range}M(x)$ the set of lifted points reachable by some output, and $\mathcal A_i(x)=h_i(x)+\operatorname{range}M_i(x)$ the reachable set of the $i$th constraint.

\begin{assumption}[Feasibility and separability]
    \label{ass:separability}\label{ass:feasibility}
    For every $x\in\mathcal X$, each constraint is feasible, i.e., $\mathcal A_i(x)\cap\mathcal K_i\ne\emptyset$ for $i=1,\ldots,p$.
    Moreover, the constraints are separable, that is,
    \begin{equation}
        \label{eq:rank-additivity}
        \operatorname{rank}M(x)=\sum_{i=1}^p\operatorname{rank}M_i(x).
    \end{equation}
\end{assumption}

\begin{remark}
    \label{rem:scope}
    The separability condition \eqref{eq:rank-additivity} concerns reachability in the lifted space so that any tuple of points from each reachable set is realizable by a common output.
    It allows rank-deficient liftings and covers constraints that share output variables.
    Moreover, the learning objective need not be separable: in Section~\ref{sec:friction}, the per-contact friction constraints accompany a wrench-matching objective that couples the contact forces.
    For a single-constraint case, \eqref{eq:rank-additivity} holds trivially for any SOC constraint, including the non-surjective input-robust CBF constraint of Section~\ref{sec:di3d}.
    % holds trivially for a single SOC constraint and whenever the constraints act on disjoint blocks of the output, as do the per-contact friction cones of Section~\ref{sec:friction} or per-actuator norm bounds.
    % It excludes constraints that compete for the same output directions, such as the coupled cone constraints of model predictive control (MPC) or sequential convex programming (SCP), which we leave to future work.
\end{remark}

The constrained network should meet three requirements:
(i) \emph{hard feasibility}, i.e., the output lies in $\mathcal S(x)$ for every $x\in\mathcal X$, not merely on average or after training;
(ii) \emph{differentiability} almost everywhere of the map from $f_\theta(x)$ to the output, so that $\theta$ can be trained end-to-end; and
(iii) \emph{no loss of expressiveness}, i.e., universal approximation over continuous functions satisfying \eqref{eq:soc}.

A natural way to meet (i) is the Euclidean projection
\begin{equation}
\mathcal{P}(f_\theta)(x)=\underset{y\in\mathcal S(x)}{\arg\min}\;\frac12\|y-f_\theta(x)\|_2^2,
\label{eq:projection-target}
\end{equation}
which is feasible and leaves a feasible $f_\theta(x)$ unchanged.
However, \eqref{eq:projection-target} has no closed form in general, so every forward pass requires a conic solve and every backward pass requires differentiating the solver's optimality conditions \citep{agrawal2019differentiable}.
We therefore seek a correction that meets (i)--(iii) without calling a conic solver in either pass.

\section{SOC-Net: Neural Networks with Hard Input-Dependent Second-Order Cone Constraints}
\label{sec:socnet}

SOC-Net lifts the backbone output to a space in which the constraints become membership in a fixed cone, corrects it there, and maps the correction back to the output space, as illustrated in Fig.~\ref{fig:pipeline}.

For the raw output $f_\theta(x)$, we call $w=M(x)f_\theta(x)+h(x)$ the lifted point.
A feasible output needs its lifted point in $\mathcal A(x)\cap\mathcal K$, which is nonempty under Assumption~\ref{ass:separability}.
The set $\mathcal A(x)\cap\mathcal K$ is then closed and convex, so the Euclidean projection onto it,
\begin{equation}
    \label{eq:lifted-correction}
    \Pi_{\mathcal A(x)\cap\mathcal K}(w)=\underset{v\in\mathcal A(x)\cap\mathcal K}{\arg\min} ~~ \frac12\|v-w\|_2^2,
\end{equation}
exists and is unique, and we define the SOC-Net correction by
\begin{equation}
    \label{eq:socnet-layer}
    \Psoc(f_{\theta})(x)=f_{\theta}(x)+M(x)^+\bigl(\Pi_{\mathcal A(x)\cap\mathcal K}(w)-w\bigr),
\end{equation}
where $M(x)^+$ is the Moore--Penrose pseudoinverse.
The lifted projection moves $w$ as little as possible within $\mathcal A(x)$, and $M(x)^+$ selects the minimum-norm output displacement realizing it.

For the $i$th constraint, let $w_i=M_i(x)f_\theta(x)+h_i(x)$ be its lifted point, so that $w=(w_1,\ldots,w_p)$.
Under Assumption~\ref{ass:separability}, the projection \eqref{eq:lifted-correction} then splits into one projection $\Pi_{\mathcal A_i(x)\cap\mathcal K_i}(w_i)$ per constraint. % (Section~\ref{sec:multiple-soc}).
For intuition, we first consider a single cone without the index $i$ in Sections~\ref{sec:surjective-soc-net} and~\ref{sec:non-surjective-socnet} and then assemble the blocks in Section~\ref{sec:multiple-soc}.
If $M(x)$ is surjective, every point in $\R^{k+1}$ is reachable and $\mathcal A(x)\cap\mathcal K=\mathcal K$ (Section~\ref{sec:surjective-soc-net}); otherwise, $\mathcal A(x)$ is a proper affine subspace and the correction must remain in the affine slice $\mathcal A(x)\cap\mathcal K$ (Section~\ref{sec:non-surjective-socnet}).

\begin{figure}[!htbp]
    \centering
    \subfigure[Surjective case\label{fig:surjective}]{%
        \includegraphics[height=0.22\linewidth]{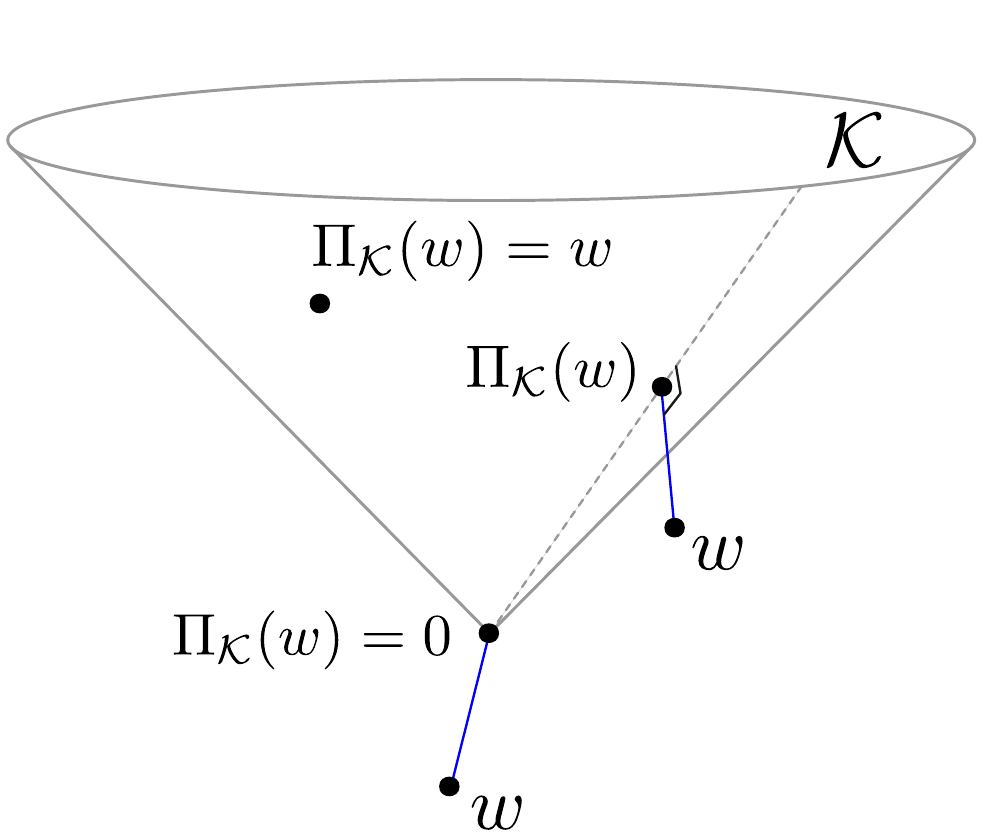}}
    \hfill
    \subfigure[Paraboloid slice\label{fig:paraboloid}]{%
        \includegraphics[height=0.22\linewidth]{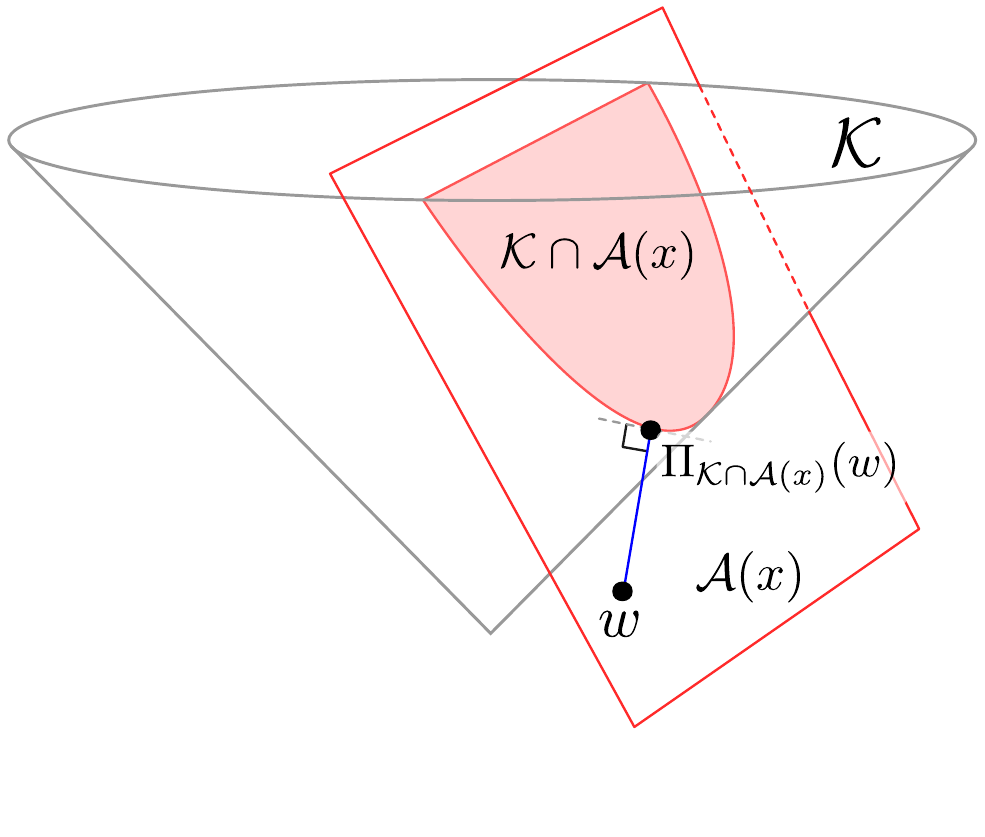}}
    \hfill
    \subfigure[Slice frame\label{fig:soc-net-diagram}]{%
        \includegraphics[height=0.22\linewidth]{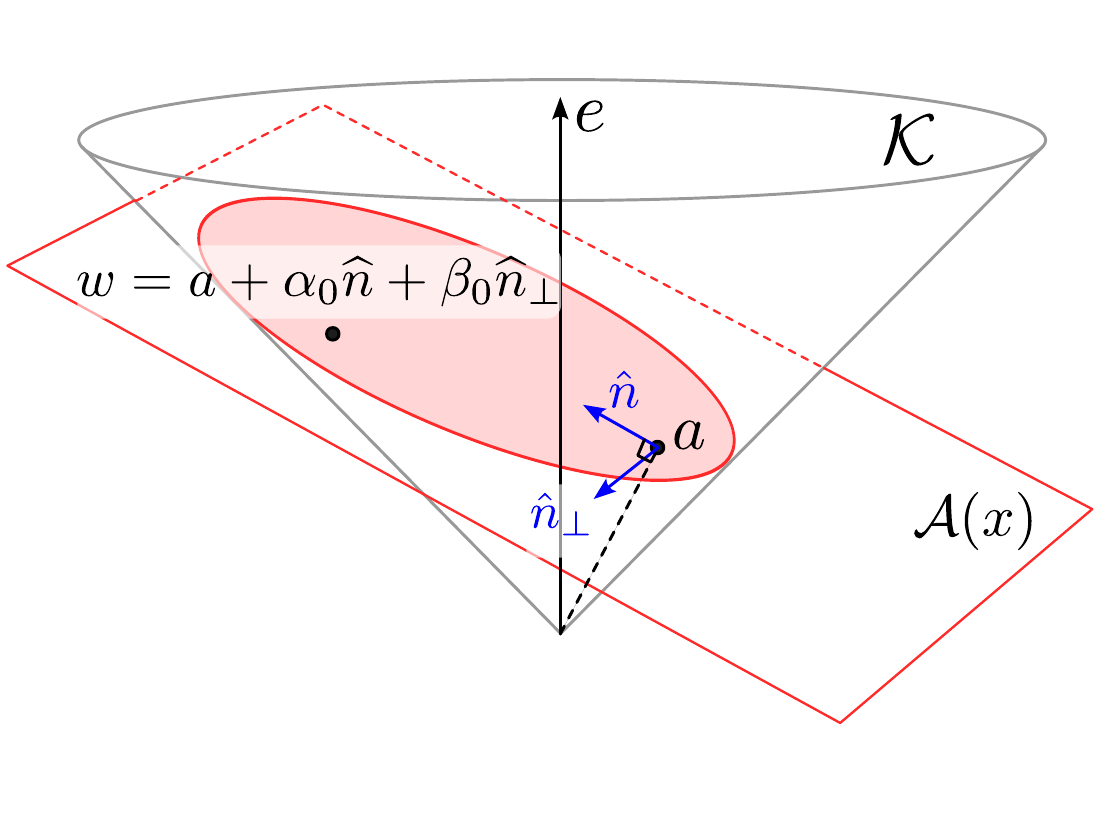}}
    \caption{Projection geometry of SOC-Net. (a) Surjective case: $w$ is projected directly onto $\mathcal K$. (b) Non-surjective case: $w\in\mathcal A(x)$ is projected within $\mathcal A(x)$ onto the slice $\mathcal A(x)\cap\mathcal K$, here a paraboloid. (c) Slice frame: $w\in\mathcal A(x)$ has coordinates $(\alpha_0,\beta_0)$ in the frame $(a,\widehat n,\widehat n_\perp)$.}
    \label{fig:soc-net}
\end{figure}

\subsection{Surjective case}
\label{sec:surjective-soc-net}

Suppose first that $M(x)$ has full row rank for every $x\in\mathcal X$, i.e., $\operatorname{rank}M(x)=k+1\le\nout$, so that $M(x)M(x)^+=I_{k+1}$.
Then every lifted point is reachable, $\mathcal A(x)=\R^{k+1}$, and \eqref{eq:socnet-layer} reduces to the standard Lorentz-cone projection:
\begin{equation}
    \label{eq:socnet-layer-surjective}
    \Psoc(f_\theta)(x) = f_\theta(x)+M(x)^+\bigl(\Pi_{\mathcal K}(w)-w\bigr).
\end{equation}
Writing $w=(w_0,\bar w)\in\R\times\R^k$, the cone projection has the well-known closed form
\begin{equation}
    \label{eq:cone-projection}
    \Pi_{\mathcal K}(w)=
    \begin{cases}
        w, & \text{if } \|\bar w\|_2\le w_0,\\[3pt]
        0_{k+1}, & \text{if } \|\bar w\|_2\le -w_0,\\[3pt]
        \dfrac{w_0+\|\bar w\|_2}{2}
        \begin{bmatrix}
            1\\
            \bar w/\|\bar w\|_2
        \end{bmatrix},
        & \text{otherwise}.
    \end{cases}
\end{equation}
The three branches correspond to $w$ lying in $\mathcal K$, in its polar cone $-\mathcal K$ (projected onto the apex), and elsewhere (projected onto the nearest boundary ray), so \eqref{eq:socnet-layer-surjective} is computed in closed form.

\subsection{Non-surjective case}
\label{sec:non-surjective-socnet}

When $M(x)$ is not surjective (i.e., $\operatorname{rank}M(x)<k+1$), $\mathcal A(x)$ is a proper affine subspace, and $\mathcal A(x)\cap\mathcal K$ is a conic section of $\mathcal K$: an ellipsoid, a paraboloid, or one sheet of a hyperboloid (Fig.~\ref{fig:soc-net}); degenerate slices are treated in Appendix~\ref{app:affine-slice}.
For $w\in\mathcal A(x)$, the cone projection $\Pi_{\mathcal K}(w)$ generally does not lie in $\mathcal A(x)$ and is therefore not reachable by any output, so SOC-Net instead projects directly onto the reachable conic section.

With $x$ fixed and dropped, let $a$ be the point of $\mathcal A$ closest to the apex, and let $\widehat n$ and $s$ be the unit direction and the length of the cone axis projected onto $\operatorname{range}M$ (Fig.~\ref{fig:soc-net-diagram}).
In this frame, the raw lifted point is $w=a+\alpha_0\widehat n+\beta_0\widehat n_\perp$, with axial coordinate $\alpha_0$, transverse distance $\beta_0\ge0$, and unit transverse direction $\widehat n_\perp$ (Appendix~\ref{app:slice-frame}).
If $w\in\mathcal K$, the projection is $w$ itself; otherwise, for a nondegenerate slice,
\begin{equation}
    \label{eq:slice-projection}
    \Pi_{\mathcal A(x)\cap\mathcal K}(w)=a+\alpha^\star\widehat n+\frac{\beta_0}{1+\lambda^\star}\widehat n_\perp.
\end{equation}
Because the slice is rotationally symmetric about $\widehat n$, the projection stays in the plane of $\widehat n$ and $\widehat n_\perp$, and its KKT conditions reduce to one monotone scalar equation in the multiplier $\lambda$, whose form depends on the sign of $\mu=1-2s^2$ (Table~\ref{tab:slice-roots}).
Each equation is a polynomial of degree at most four after clearing denominators, regardless of $k$ and $\operatorname{rank}M(x)$, and the implementation solves it by a safeguarded Newton method.
The derivation, degenerate slices, and a feasibility-preserving finite-accuracy reconstruction are in Appendix~\ref{app:affine-slice}.

\begin{table}[!htbp]
    \centering
    \small
    \renewcommand{\arraystretch}{1.3}
    \caption{Scalar equations for the projection onto a nondegenerate slice, with $\tau=e^\top a$, $\alpha_c=2\tau s/\mu$, and $q_c=\|a\|_2^2-2\tau^2/\mu$ for $\mu\ne0$: $\lambda^\star>0$ is the unique positive root of the (unsquared) scalar equation, and $\alpha^\star$ is the corresponding axial coordinate in \eqref{eq:slice-projection}.}
    \label{tab:slice-roots}
    \begin{tabular}{@{}lll@{}}
        \toprule
        Slice & Scalar equation for $\lambda^\star>0$ & Axial coordinate $\alpha^\star$ \\
        \midrule
        Ellipsoid ($\mu>0$)
        & $\displaystyle \frac{\beta_0^2}{(1+\lambda)^2}+\frac{\mu(\alpha_0-\alpha_c)^2}{(1+\mu\lambda)^2}+q_c=0$
        & $\displaystyle \alpha_c+\frac{\alpha_0-\alpha_c}{1+\mu\lambda^\star}$ \\
        Paraboloid ($\mu=0$)
        & $\displaystyle \frac{\beta_0^2}{(1+\lambda)^2}+\|a\|_2^2-2\tau^2-4\tau s\alpha_0-8\tau^2s^2\lambda=0$
        & $\displaystyle \alpha_0+2\tau s\lambda^\star$ \\
        Hyperboloid ($\mu<0$)
        & $\displaystyle (1+\mu\lambda)\sqrt{\frac{\beta_0^2}{(1+\lambda)^2}+q_c}-\sqrt{-\mu}\,(\alpha_0-\alpha_c)=0$
        & $\displaystyle \alpha_c+\sqrt{\frac{\beta_0^2/(1+\lambda^\star)^2+q_c}{-\mu}}$ \\
        \bottomrule
    \end{tabular}
\end{table}

\subsection{Multiple SOC constraints}
\label{sec:multiple-soc}

We now combine the single-cone projections of Sections~\ref{sec:surjective-soc-net} and~\ref{sec:non-surjective-socnet} into a projection layer for multiple SOC constraints.
This rests on the fact that, under Assumption~\ref{ass:separability}, the reachable subspace is the product of the per-constraint reachable subspaces,
\begin{equation}
    \label{eq:range-product}
    \operatorname{range}M(x)=\prod_{i=1}^p\operatorname{range}M_i(x).
\end{equation}
The inclusion $\subseteq$ always holds, because the $i$th block of $M(x)y$ is $M_i(x)y\in\operatorname{range}M_i(x)$ for every output $y$.
The reverse inclusion follows from the separability condition \eqref{eq:rank-additivity}, which gives both subspaces the dimension $\sum_i\operatorname{rank}M_i(x)$. Adding the offset $h(x)$ to both sides of \eqref{eq:range-product} gives
\begin{equation}
    \label{eq:slice-product}
    \mathcal A(x)=\prod_{i=1}^p\mathcal A_i(x),
    \qquad
    \mathcal A(x)\cap\mathcal K=\prod_{i=1}^p\bigl(\mathcal A_i(x)\cap\mathcal K_i\bigr).
\end{equation}
Any tuple of individually reachable lifted points is thus reached by a single output.

With this factorization, the SOC-Net layer \eqref{eq:socnet-layer} for the stacked lifting becomes
\begin{equation}
    \label{eq:multi-socnet-layer}
    \Psoc(f_\theta)(x)=f_\theta(x)+M(x)^+
    \begin{bmatrix}
        \Pi_{\mathcal A_1(x)\cap\mathcal K_1}(w_1)-w_1\\
        \vdots\\
        \Pi_{\mathcal A_p(x)\cap\mathcal K_p}(w_p)-w_p
    \end{bmatrix}.
\end{equation}
Surjective and non-surjective constraints can therefore coexist in one layer.
A block with surjective $M_i(x)$ has $\mathcal A_i(x)=\R^{k_i+1}$ and uses the closed-form cone projection $\Pi_{\mathcal K_i}(w_i)$ in \eqref{eq:cone-projection}, while every other block uses the slice projection \eqref{eq:slice-projection}.

\subsection{Optimization view}
\label{sec:optimization-view}

SOC-Net does not compute the Euclidean projection \eqref{eq:projection-target}, which has no closed form in general.
Its output is instead the exact solution of a closely related nested projection problem, analogous to HardNet-Aff \citep[Prop.~6]{min2024hardnet}; see Appendix~\ref{app:variational-proof} for a proof.

\begin{proposition}[Optimization characterization]
    \label{prop:variational}
    Let Assumption~\ref{ass:feasibility} hold and fix $x\in\mathcal X$.
    Then the SOC-Net output \eqref{eq:socnet-layer} is the unique solution of the bilevel problem
    \begin{equation}
        \label{eq:bilevel}
        \Psoc(f_\theta)(x)=\underset{z\in\R^{\nout}}{\arg\min}\ \|z-f_\theta(x)\|_2
        \quad\text{s.t.}\quad
        z\in\underset{y\in\mathcal S(x)}{\arg\min}\ \bigl\|M(x)\bigl(y-f_\theta(x)\bigr)\bigr\|_2 .
    \end{equation}
\end{proposition}

Thus, \eqref{eq:bilevel} and \eqref{eq:projection-target} share the feasible set $\mathcal S(x)$ and differ only in how the correction $y-f_\theta(x)$ is measured: \eqref{eq:projection-target} measures it in the output space, whereas the inner problem of \eqref{eq:bilevel} measures it in the lifted space, which reduces it to the block-wise cone or slice projections of Sections~\ref{sec:surjective-soc-net}--\ref{sec:multiple-soc}.
When $M(x)$ has a nontrivial null space, the inner problem has infinitely many solutions, which differ by elements of $\ker M(x)$, and SOC-Net chooses the one closest to $f_\theta(x)$ in Euclidean distance.

\section{Feasibility, differentiability, and universal approximation}
\label{sec:socnet-properties}

We now show that SOC-Net meets (i)--(iii) of Section~\ref{prob_form}; proofs are in Appendices~\ref{app:slice-recovery}--\ref{app:uat-proof}.

\begin{theorem}[Exact feasibility]
    \label{thm:slice-layer-properties}
    Let Assumption~\ref{ass:feasibility} hold.
    For every $x\in\mathcal X$, the SOC-Net output \eqref{eq:socnet-layer} is well defined and satisfies
    \begin{equation}
        \label{eq:slice-layer-feasibility}
        \Psoc(f_\theta)(x)\in\mathcal S(x),
        \quad\text{i.e.,}\quad
        M(x)\Psoc(f_\theta)(x)+h(x)=\Pi_{\mathcal A(x)\cap\mathcal K}(w)\in\mathcal K.
    \end{equation}
    Moreover, if $f_\theta(x)\in\mathcal S(x)$, then $\Psoc(f_\theta)(x)=f_\theta(x)$.
\end{theorem}

Feasibility thus holds for every input and parameter $\theta$, independently of training.
It also does not require solving the scalar equation in Table~\ref{tab:slice-roots} exactly: SOC-Net maps any approximate root to a point on the boundary of the slice (Appendix~\ref{app:affine-slice}), so stopping the Newton iteration early makes the correction less accurate but never infeasible.

\begin{theorem}[Differentiability]
    \label{thm:differentiability}
    Fix $x$ with $\operatorname{rank}M(x)=r\ge1$.
    Let $\sigma_{\max}(M(x))$ and $\sigma_r(M(x))$ be the largest and smallest nonzero singular values of $M(x)$, and define
    \begin{equation}
        \label{eq:positive-condition-number}
        \kappa_+(M(x))=\sigma_{\max}(M(x))/\sigma_r(M(x))\ge1.
    \end{equation}
    The map $f_\theta(x)\mapsto\Psoc(f_\theta)(x)$ is $\kappa_+(M(x))$-Lipschitz and hence differentiable almost everywhere.
    Wherever $\Pi_{\mathcal A(x)\cap\mathcal K}$ is differentiable at $w$, the Jacobian is
    \begin{equation}
        \label{eq:slice-layer-jacobian}
        D_{f_\theta(x)}\Psoc(f_\theta)(x)=I+M(x)^+\bigl\{D_w\Pi_{\mathcal A(x)\cap\mathcal K}(w)-I\bigr\}M(x).
    \end{equation}
\end{theorem}

Theorem~\ref{thm:differentiability} meets requirement (ii): the gradient of any training loss passes through the correction to reach the backbone, so $\theta$ can be trained end-to-end by backpropagation.
The backward pass is also cheap, since \eqref{eq:slice-layer-jacobian} requires only $D_w\Pi_{\mathcal A(x)\cap\mathcal K}(w)$.
Block by block, this is the Jacobian of the closed form \eqref{eq:cone-projection} in the surjective case, and in the non-surjective case it follows from the implicit function theorem applied to the scalar equation in Table~\ref{tab:slice-roots}, without unrolling the Newton iteration (Appendix~\ref{ssec:implicit-differentiation}).

\begin{theorem}[Universal approximation]
    \label{thm:uat}
    Let $\mathcal X$ be compact, let $M$ and $h$ be continuous, and let $\operatorname{rank}M(x)=r\ge1$ be constant on $\mathcal X$.
    Suppose that the moving projector $(x,w)\mapsto\Pi_{\mathcal A(x)\cap\mathcal K}(w)$, $w\in\mathcal A(x)$, is continuous.
    Let $\mathcal F\subset C(\mathcal X;\R^{\nout})$ be a class of backbone networks that universally approximates (or is dense in) $C(\mathcal X;\R^{\nout})$ in the uniform norm, and let $\overline\kappa=\max_{x\in\mathcal X}\kappa_+(M(x))$.
    Then, for every $f\in\mathcal F$, $\Psoc(f)$ is continuous and feasible, and for every continuous $g$ with $g(x)\in\mathcal S(x)$ for all $x\in\mathcal X$,
    \begin{equation}
        \label{eq:uat-bound}
        \|\Psoc(f)-g\|_\infty\le\overline\kappa\,\|f-g\|_\infty.
    \end{equation}
    Consequently, $\Psoc(\mathcal F)$ universally approximates every such $g$.
\end{theorem}

Theorem~\ref{thm:uat} meets requirement (iii), since every continuous feasible target $g$ is uniformly approximated by some $f\in\mathcal F$ and Theorem~\ref{thm:uat} shows that the correction preserves this approximation up to the factor $\overline\kappa$.
This factor follows from two facts: the correction leaves the feasible target $g$ unchanged (Theorem~\ref{thm:slice-layer-properties}), and it is $\overline\kappa$-Lipschitz (Theorem~\ref{thm:differentiability}).
Therefore, the projection layer $\Psoc$ does not restrict which feasible maps the network can represent.

% ==============

\section{Experiments}
\label{sec:exp}

We evaluate SOC-Net on two control tasks, one for each case of Section~\ref{sec:socnet}: contact force allocation, where several separable constraints each have a surjective lifting (Section~\ref{sec:multiple-soc}), and safe control of a 3-D aircraft under actuation uncertainty, where a single constraint has a non-surjective lifting.
Both use float64, with learned methods on an RTX~5090 GPU and conic solvers on one CPU core (Appendices~\ref{app:friction} and~\ref{app:di3d}).

\subsection{Contact force allocation for grasping with friction constraints}
\label{sec:friction}

We consider a rigid object supported at four coplanar point contacts with parallel upward normals (Fig.~\ref{fig:friction} in Appendix~\ref{app:friction}).
The task is to allocate contact forces $y=(f_1,\ldots,f_4)\in\R^{12}$, $f_i=(f_{ix},f_{iy},f_{iz})$, that reproduce a given desired wrench $x=(F_{\mathrm{des}},\tau_{\mathrm{des}})\in\R^6$ while staying inside the friction cones to prevent slip.
This allocation can be posed as an optimization problem with the friction cones as hard constraints and the wrench error as the objective,
\begin{equation}
    \min_{y\in\R^{12}}\ \|Gy-x\|_2^2
    \quad\text{s.t.}\quad
    \|(f_{ix},f_{iy})\|_2\le\mu_{\mathrm f} f_{iz},\quad i=1,\ldots,4,
    \label{eq:friction-task-main}
\end{equation}
where the grasp map $G\in\R^{6\times12}$ gives $Gy=(\sum_i f_i,\sum_i r_i\times f_i)$ for contact positions $r_i$, and $\mu_{\mathrm f}>0$ is the friction coefficient.
Each friction cone is one constraint of \eqref{eq:soc} with $h_i=0$ and the lifting $M_i\in\R^{3\times12}$ given by $M_iy=(\mu_{\mathrm f}f_{iz},f_{ix},f_{iy})$.
Since $M=[M_1;\ldots;M_4]$ satisfies Assumption~\ref{ass:separability} and every $M_i$ is surjective, we use the layer \eqref{eq:multi-socnet-layer} with the closed-form cone projection \eqref{eq:cone-projection} at each contact.
We compare SOC-Net with a raw network and a soft-penalty network on the same backbone, and with per-instance solutions of \eqref{eq:friction-task-main} by the SOCP solvers ECOS and Clarabel.

The penalty network still violates a friction cone on $0.19\%$ of the test inputs, whereas SOC-Net's largest violation is $2.9\times10^{-16}\,\mathrm N$, which is floating-point round-off (Table~\ref{tab:friction}).
Because its correction is in closed form, SOC-Net also processes the test set in $0.179\,\mathrm{ms}$, about $1.1\times10^{4}$ and $1.4\times10^{4}$ times faster than ECOS and Clarabel.

\begin{table}[!htbp]
\centering
\caption{Contact force allocation on 4,096 test wrenches, mean over five seeds.
The friction violation of an output is its largest cone violation over the four contacts, and its rate counts inputs above $10^{-6}\,\mathrm N$ (Appendix~\ref{app:friction}).}
\label{tab:friction}
\footnotesize
\setlength{\tabcolsep}{4pt}
\begin{tabular}{lccccc}
\toprule
Method & Wrench MSE & \shortstack{Max (mean) friction\\viol.\ (N)} & \shortstack{Friction viol.\\rate (\%)} & \shortstack{$T_{\mathrm{test}}$\\(ms)} & \shortstack{$T_{\mathrm{train}}$\\(s)} \\
\midrule
Raw NN & $6.47\times10^{-4}$ & $0.508\;(4.62\times10^{-2})$ & $31.46$ & $0.117$ & $2.27$ \\
Penalty NN ($\lambda=100$) & $2.333\times10^{-3}$ & $0.0423\;(3.09\times10^{-5})$ & $0.1855$ & $0.117$ & $3.31$ \\
\textbf{SOC-Net (Ours)} & \textbf{$2.585\times10^{-3}$} & \textbf{$2.9\times10^{-16}\;(5.4\times10^{-17})$} & $0$ & $0.179$ & $4.60$ \\
\midrule
SOCP (Clarabel) & $9.41\times10^{-19}$ & $0\;(0)$ & $0$ & $2{,}488$ & -- \\
SOCP (ECOS) & $3.40\times10^{-12}$ & $0\;(0)$ & $0$ & $2{,}052$ & -- \\
\bottomrule
\end{tabular}
\end{table}

\subsection{3-D aircraft obstacle avoidance under actuation uncertainty}
\label{sec:di3d}
\suppressfloats[t]

\begin{figure*}[!htb]
\centering
\begin{minipage}[b]{0.46\linewidth}\centering
\includegraphics[width=\linewidth,trim=0 3 0 24,clip]{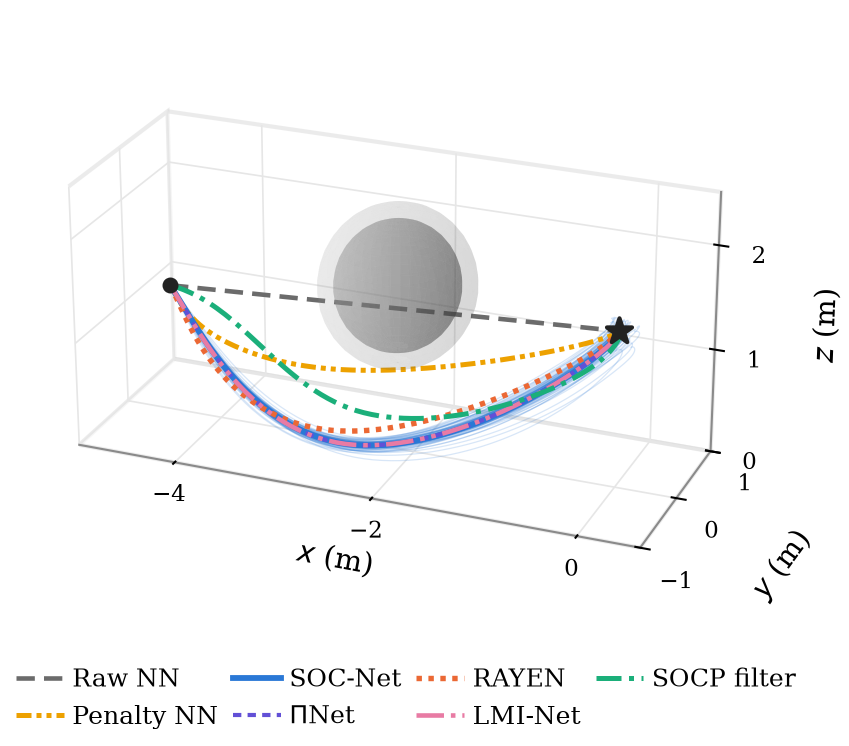}
\end{minipage}\hspace{0.06\linewidth}
\begin{minipage}[b]{0.46\linewidth}\centering
\includegraphics[width=\linewidth]{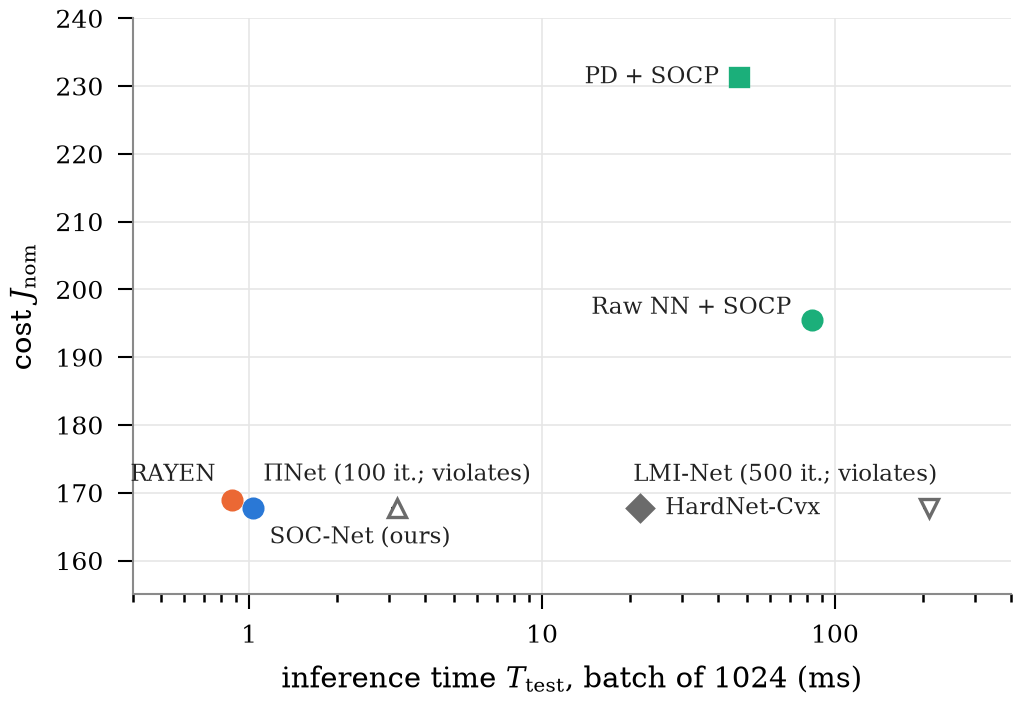}
\end{minipage}
\vspace{-6pt}
\caption{(a) Nominal trajectories from one test state (``SOCP filter'' is Raw NN\,+\,SOCP filter) and SOC-Net's random-uncertainty rollouts (thin blue). (b) $J_{\mathrm{nom}}$ against $T_{\mathrm{test}}$ over five seeds.}
\label{fig:di3d}
\end{figure*}

We consider a simplified 3-D aircraft with state $x=(p,v)\in\R^6$ of position and velocity, and commanded acceleration $u\in\R^3$.
Its dynamics are given by $\dot p=v$, $\dot v=(I+\Delta)u+\delta-ge_z$ with the unknown actuator gain error $\Delta\in\R^{3\times3}$, $\|\Delta\|_2\le\varepsilon_g=0.2$ and the disturbance $\delta\in\R^3$, $\|\delta\|_2\le\varepsilon_d=0.5$.
Starting at rest, the vehicle must reach a goal $p^\star$ that lies behind a spherical obstacle (Fig.~\ref{fig:di3d}a).

We train a residual policy on top of a PD controller $u_{\mathrm{PD}}(x)$.
At each control instant $k$, the command $u_k = u_{\mathrm{PD}}(x_k)+\pi_{\mathrm{NN}}(x_k)$ is applied and held over a step of length $t_{\mathrm s}$.
The neural network parameters of $\pi_{\mathrm{NN}}$ are trained end-to-end through the layer to minimize the tracking cost of the nominal rollout with $\Delta=0$ and $\delta=0$,
\begin{equation}
J=\sum_{k=0}^{N-1}t_{\mathrm s}\bigl(\|p_k-p^\star\|_{Q_p}^2+\|v_k\|_{Q_v}^2+\|u_k-ge_z\|_{R}^2\bigr),
\label{eq:di3d-cost}
\end{equation}
where $N$ is the horizon and $\|z\|_Q^2=z^\top Qz$ with weights $Q_p,Q_v,R\succ0$.

The applied command must also avoid the obstacle under every admissible uncertainty, which we encode with a control barrier function (CBF) $B(x)$ built from the obstacle distance.
If every applied input satisfies $\dot B+k_BB\ge0$ for a gain $k_B>0$, the set $\{B\ge0\}$ is forward invariant, and the vehicle never enters the obstacle \citep{ames2017control}.
Requiring the CBF condition for every admissible $(\Delta,\delta)$ yields the input-robust CBF constraint \citep{taylor2021towards}
\begin{equation}
\varepsilon_g\|c(x)\|_2\|u\|_2\le c(x)^\top u+d(x),
\label{eq:di3d-robust-soc}
\end{equation}
which is an SOC constraint in $u$, with $c(x)$ and $d(x)$ derived in Appendix~\ref{app:di3d}.
We therefore implement $\pi_{\mathrm{NN}}$ as an SOC-Net, whose lifting here is the non-surjective case introduced in Section~\ref{sec:non-surjective-socnet}.

Each policy is evaluated from $100$ initial states under $102$ uncertainty scenarios (the nominal one, a worst case that pushes the vehicle toward the obstacle at every step, and $100$ random draws), and $J_{\mathrm{nom}}$ and $J_{\mathrm{wc}}$ denote the cost \eqref{eq:di3d-cost} of the nominal and worst-case rollouts.
Baselines are a raw and a penalty network, the hard-constrained layers RAYEN, $\Pi$Net, HardNet-Cvx, and LMI-Net, and SOCP safety filters that project the raw network's or the PD controller's command onto \eqref{eq:di3d-robust-soc} online.

SOC-Net satisfies \eqref{eq:di3d-robust-soc} in every rollout, whereas $\Pi$Net, LMI-Net, and the penalty network violate it in $68\%$, $23\%$, and $16\%$ of rollouts (Table~\ref{tab:di3d}).
Because SOC-Net projects onto the feasible set instead of scaling toward an interior point as RAYEN does, it attains a lower cost than RAYEN and matches HardNet-Cvx, which computes the Euclidean projection with a conic solver.
It does so in $1.03\,\mathrm{ms}$ per batch, $21\times$ and $81\times$ faster than HardNet-Cvx and the SOCP filter (Fig.~\ref{fig:di3d}b).

\begin{table*}[!t]
\centering
\caption{3-D obstacle avoidance, mean $\pm$ std over five seeds.
CBF violation is the largest violation of \eqref{eq:di3d-robust-soc} along a rollout, at relative tolerance $10^{-9}$ for the rate. $^\dagger$Our implementation.}
\label{tab:di3d}
\footnotesize
\setlength{\tabcolsep}{4pt}
\begin{tabular}{lcccccc}
\toprule
Method & $J_{\mathrm{nom}}$ & $J_{\mathrm{wc}}$ & \shortstack{Max (mean)\\CBF viol.} & \shortstack{CBF viol.\\rate (\%)} & \shortstack{$T_{\mathrm{test}}$\\(ms)} & \shortstack{$T_{\mathrm{train}}$\\(s)} \\
\midrule
Raw NN & 127.1 & 123.4 & $3.7\times10^{2}\;(3.2\times10^{2})$ & 100 & 0.31 & 1422 \\
Penalty NN ($\lambda=10^3$) & $187.2\pm9.0$ & 172.4 & $1.6\times10^{1}\;(3.9\times10^{-1})$ & $15.6\pm8.6$ & 0.31 & 1617 \\
\midrule
\textbf{SOC-Net (Ours)} & 167.7 & 160.7 & $2.3\times10^{-13}\;(4.9\times10^{-14})$ & \textbf{0} & 1.03 & \textbf{1873} \\
RAYEN & $169.0\pm0.5$ & 168.9 & $3.0\times10^{-13}\;(8.7\times10^{-14})$ & 0 & \textbf{0.87} & 2234 \\
$\Pi$Net (100 it.) & 167.7 & 160.6 & $5.8\times10^{-3}\;(1.4\times10^{-5})$ & $67.8\pm9.6$ & 3.2 & 3401 \\
HardNet-Cvx & 167.7 & 160.0 & $2.1\times10^{-8}\;(5.9\times10^{-10})$ & 0 & 21.6 & 5391 \\
LMI-Net$^\dagger$ (500 it.) & 167.7 & 168.3 & $2.2\times10^{-5}\;(1.1\times10^{-6})$ & $22.7\pm2.1$ & 209 & 61332 \\
\midrule
Raw NN + SOCP filter & $195.4\pm0.8$ & 183.1 & $2.7\times10^{-8}\;(7.8\times10^{-10})$ & 0 & 83.3 & -- \\
PD + SOCP filter & 231.3 & 214.3 & $1.0\times10^{-8}\;(2.0\times10^{-10})$ & 0 & 47.2 & -- \\
\bottomrule
\end{tabular}
\end{table*}

\FloatBarrier

\section{Conclusion}
\label{sec:conclusion}

SOC-Net is a novel neural network architecture whose output satisfies hard input-dependent SOC constraints for every input.
Its projection layer is differentiable almost everywhere with an explicit Jacobian, so the whole network can be trained end-to-end while retaining universal approximation of continuous feasible maps.
This makes it suited to applications where prior knowledge must hold exactly for every input and a conic solve per input is too slow, such as the control problems considered here.
It would be interesting to extend SOC-Net beyond the separable setting of Assumption~\ref{ass:separability} to non-separable SOC constraints, which would make it applicable to problems such as trajectory optimization where many coupled cone constraints act on the same decision variables.

\subsection*{Generative AI Use}
Generative AI tools were used for drafting parts of the manuscript and implementing parts of the experiments. The authors reviewed and verified all AI-assisted content and take full responsibility for the final content of the paper.

\acks{We sincerely thank Andrea Goertzen and Sunbochen Tang for providing feedback during the early stages of this work. This work was supported in part by the National Research Foundation of Korea (NRF) grant from the Korean government (MSIT) (RS-2026-25504174), by the U.S. National Science Foundation under CAREER Award No.2544396, and by the MIT-Amazon Science Hub.}

\bibliography{reference}

%%%%%%%%%%%%%%%%%%%%%%%%%%%%%%%%%%%%%%%%%%%%%%%%%%%%%%%%%%%%%%%%%%%%%%%%%%%%%%%%

\appendix

\section{Derivation of the affine-slice projection}
\label{app:affine-slice}

Throughout this section, $x$ is fixed and dropped from the notation, and $M$, $h$, $\mathcal K$, and $\mathcal A$ denote the quantities of a single constraint, as in Sections~\ref{sec:surjective-soc-net}--\ref{sec:non-surjective-socnet}.

\subsection{Slice frame and conic-section form}
\label{app:slice-frame}

Since $MM^+$ is the orthogonal projector onto $\operatorname{range}M$, the frame of Section~\ref{sec:non-surjective-socnet} is
\begin{equation*}
a=h-MM^+h, \qquad s=\|MM^+e\|_2, \qquad \widehat n=\frac{MM^+e}{s},
\end{equation*}
and every $v\in\mathcal A$ has axial coordinate $\alpha(v)=\widehat n^\top(v-a)$ and transverse distance $\beta(v)=\|v-a-\alpha(v)\widehat n\|_2$.
In particular, the raw lifted point $w=Mf_\theta+h$ decomposes as
\begin{equation}
    \label{eq:slice-decomposition}
    w=a+\alpha_0\widehat n+\beta_0\widehat n_\perp,
    \qquad
    \alpha_0=\alpha(w),
    \quad
    \beta_0=\beta(w),
    \quad
    \widehat n_\perp=\frac{w-a-\alpha_0\widehat n}{\beta_0}.
\end{equation}
When $\beta_0=0$, the transverse direction is immaterial and every term containing $\beta_0\widehat n_\perp$ is understood as zero.

Because $a\perp\operatorname{range}M$ and $e^\top\xi=s\,\widehat n^\top\xi$ for every $\xi\in\operatorname{range}M$, every $v\in\mathcal A$ satisfies $\|v\|_2^2=\|a\|_2^2+\alpha(v)^2+\beta(v)^2$ and $e^\top v=\tau+s\alpha(v)$ with $\tau=e^\top a$.
Writing $\mathcal K=\{v:\ \|v\|_2^2-2(e^\top v)^2\le0,\ e^\top v\ge0\}$ and substituting these identities gives
\begin{equation}
\begin{gathered}
\mathcal A\cap\mathcal K=\bigl\{v\in\mathcal A:\ Q(\alpha(v),\beta(v))\le0,\ \tau+s\alpha(v)\ge0\bigr\},\\
Q(\alpha,\beta)=\beta^2+\mu\alpha^2-4\tau s\alpha+\|a\|_2^2-2\tau^2, \qquad \mu=1-2s^2,
\end{gathered}
\label{eq:canonical-slice}
\end{equation}
so the slice is rotationally symmetric about the tilted axis $\{a+\alpha\widehat n:\ \alpha\in\R\}$.
For $\mu\ne0$, completing the square gives $Q(\alpha,\beta)=\beta^2+\mu(\alpha-\alpha_c)^2+q_c$ with the constants of Table~\ref{tab:slice-roots}.
The slice is therefore an ellipsoid for $\mu>0$, nonempty if and only if $q_c\le0$; a paraboloid for $\mu=0$, where Assumption~\ref{ass:feasibility} gives $\tau\ge0$ with $\tau=0$ only if $a=0$; and, for $\mu<0$, the sheet
\begin{equation}
\mathcal A\cap\mathcal K=\Bigl\{v\in\mathcal A:\ \alpha(v)-\alpha_c\ge\sqrt{(\beta(v)^2+q_c)/(-\mu)}\Bigr\}
\label{eq:hyperboloid-sheet}
\end{equation}
of a two-sheeted hyperboloid, where $q_c\ge\|a\|_2^2\ge0$.
For $\mu\ge0$ and $a\ne0$, the sets $\mathcal A\cap\mathcal K$ and $\mathcal A\cap(-\mathcal K)$ are disjoint, closed, and have the convex union $\{Q\le0\}$, so Assumption~\ref{ass:feasibility} gives $\{Q\le0\}=\mathcal A\cap\mathcal K$ and the sign condition in \eqref{eq:canonical-slice} is redundant.

\subsection{Scalar equations}
\label{app:slice-kkt}

By rotational symmetry and uniqueness, the projection of $w$ lies in the plane spanned by $\widehat n$ and $\widehat n_\perp$, so \eqref{eq:lifted-correction} reduces to
\begin{equation}
\min_{\alpha,\beta}\ \frac12\left\{(\alpha-\alpha_0)^2+(\beta-\beta_0)^2\right\}
\quad\mathrm{s.t.}\quad Q(\alpha,\beta)\le0,\ \ \tau+s\alpha\ge0.
\label{eq:reduced-slice-projection}
\end{equation}
We call the slice \emph{nondegenerate} if $s>0$ and, in addition, $q_c\ne0$ when $\mu\ne0$ or $\tau\ne0$ when $\mu=0$; then $a\ne0$ and $\nabla Q\ne0$ on $\{Q=0\}$.
For $w\notin\mathcal K$, only $Q\le0$ is active at the solution, and KKT stationarity with multiplier $\lambda^\star>0$ gives
\begin{equation}
\beta^\star=\frac{\beta_0}{1+\lambda^\star}, \qquad (1+\mu\lambda^\star)\alpha^\star =\alpha_0+2\tau s\lambda^\star,
\label{eq:common-reconstruction}
\end{equation}
where the second relation reads $(1+\mu\lambda^\star)(\alpha^\star-\alpha_c)=\alpha_0-\alpha_c$ for $\mu\ne0$.
Substituting \eqref{eq:common-reconstruction} into $Q=0$, or for $\mu<0$ into the boundary of \eqref{eq:hyperboloid-sheet}, yields the equations $F_E$, $F_P$, and $H$ of Table~\ref{tab:slice-roots}, which give \eqref{eq:slice-projection}.
Each has exactly one positive root: it is positive at $\lambda=0$ because $w\notin\mathcal K$, tends to $q_c<0$, $-\infty$, and $-\infty$, respectively, as $\lambda\to\infty$, and is strictly decreasing,
\begin{align}
F_E'(\lambda) &=-\frac{2\beta_0^2}{(1+\lambda)^3} -\frac{2\mu^2(\alpha_0-\alpha_c)^2}{(1+\mu\lambda)^3}<0,
\label{eq:ellipsoid-monotonicity}\\
F_P'(\lambda) &=-\frac{2\beta_0^2}{(1+\lambda)^3} -8\tau^2s^2<0,
\label{eq:paraboloid-monotonicity}\\
H'(\lambda) &=-\frac{-\mu q_c+(1-\mu)\beta_0^2/(1+\lambda)^3} {\sqrt{\beta_0^2/(1+\lambda)^2+q_c}}<0,
\label{eq:hyperboloid-monotonicity}
\end{align}
where the last inequality uses $-\mu q_c>0$.
Clearing denominators, after squaring for $H$, gives a polynomial of degree at most four.

\paragraph{Finite-accuracy reconstruction.}
Because the exact projection lies on the boundary of the slice, substituting an inexact root $\widetilde\lambda$ into \eqref{eq:common-reconstruction} does not by itself preserve feasibility.
SOC-Net instead sets $\overline\beta=\beta_0/(1+\widetilde\lambda)$ and maps it to the boundary:
\begin{equation}
\begin{array}{ll}
\mu>0:
&\displaystyle
\overline\alpha=\alpha_c+\frac{\alpha_0-\alpha_c}{1+\mu\widetilde\lambda},\qquad
\zeta=\sqrt{\frac{-q_c}{\mu(\overline\alpha-\alpha_c)^2+\overline\beta^2}},\\[10pt]
&\displaystyle
\widetilde\alpha=\alpha_c+\zeta(\overline\alpha-\alpha_c),\qquad
\widetilde\beta=\zeta\overline\beta,\\[10pt]
\mu=0:
&\displaystyle
\widetilde\alpha=\frac{\overline\beta^2+\|a\|_2^2-2\tau^2}{4\tau s},\qquad
\widetilde\beta=\overline\beta,\\[10pt]
\mu<0:
&\displaystyle
\widetilde\alpha=\alpha_c+\sqrt{\frac{\overline\beta^2+q_c}{-\mu}},\qquad
\widetilde\beta=\overline\beta.
\end{array}
\label{eq:finite-accuracy-reconstruction}
\end{equation}
For every $\widetilde\lambda\ge0$, the point $a+\widetilde\alpha\widehat n+\widetilde\beta\widehat n_\perp$ lies in $\mathcal A\cap\partial\mathcal K$ in exact arithmetic, so stopping the Newton iteration early affects the accuracy of the projection but not feasibility; in floating point, feasibility holds up to round-off.

\paragraph{Degenerate slices.}
The remaining cases have explicit projections.
\begin{itemize}
\item $s=0$: the slice is perpendicular to the axis, the height $e^\top v=\tau$ is constant on $\mathcal A$, and the slice is the ball $\{a+\xi:\ \xi\in\operatorname{range}M,\ \|\xi\|_2^2\le2\tau^2-\|a\|_2^2\}$ with $\tau\ge0$. Its projection is $a+\min\bigl\{1,\sqrt{2\tau^2-\|a\|_2^2}/\|w-a\|_2\bigr\}(w-a)$.
\item $\mu>0$ and $q_c=0$: the slice is the single point $a+\alpha_c\widehat n$.
\item $\mu=0$ and $\tau=0$: then $a=0$, the slice is the ray $\{\alpha\widehat n:\ \alpha\ge0\}$, and the projection is $\max\{\alpha_0,0\}\,\widehat n$.
\item $\mu<0$ and $q_c=0$: then $a=0$, and the slice is the circular cone $\{v\in\operatorname{range}M:\ \beta(v)\le c\,\alpha(v)\}$ with $c=\sqrt{-\mu}$. Its projection is $w$ if $\beta_0\le c\,\alpha_0$, the apex if $\alpha_0+c\beta_0\le0$, and otherwise $\frac{\alpha_0+c\beta_0}{1+c^2}(\widehat n+c\widehat n_\perp)$, which reduces to \eqref{eq:cone-projection} when $c=1$.
\end{itemize}
These branches also include every projection onto the cone apex.

\section{Proof of Theorem~\ref{thm:slice-layer-properties}}
\label{app:slice-recovery}

Let $v^\star=\Pi_{\mathcal A(x)\cap\mathcal K}(w)$.
The projection exists and is unique because $\mathcal A(x)\cap\mathcal K$ is nonempty, closed, and convex.
Since $v^\star,w\in\mathcal A(x)$, their difference belongs to $\operatorname{range}M(x)$.
Therefore,
\begin{equation*}
    M(x)\Psoc(f_\theta)(x)+h(x)=w+M(x)M(x)^+(v^\star-w)=v^\star\in\mathcal K.
\end{equation*}
If $f_\theta(x)$ is feasible, then $w\in\mathcal A(x)\cap\mathcal K$ and hence $v^\star=w$.
This proves the identity property.
Applying the identity property to a corrected output proves idempotence.
Finally,
\begin{equation*}
    \bigl(I-M(x)^+M(x)\bigr)\Psoc(f_\theta)(x)=\bigl(I-M(x)^+M(x)\bigr)f_\theta(x),
\end{equation*}
because the correction belongs to $\operatorname{range}M(x)^\top$.
\hfill$\square$

\section{Proof of Proposition~\ref{prop:variational}}
\label{app:variational-proof}

Fix $x$ and drop it from the notation; write $f=f_\theta(x)$, $w=Mf+h$, and $v^\star=\Pi_{\mathcal A\cap\mathcal K}(w)$.
As $y$ ranges over $\R^{\nout}$, the lifted point $v=My+h$ ranges over $\mathcal A$, and $\|M(y-f)\|_2=\|v-w\|_2$; moreover, $y\in\mathcal S$ if and only if $v\in\mathcal K$.
Hence the inner problem has optimal value $\min_{v\in\mathcal A\cap\mathcal K}\|v-w\|_2$, attained exactly by the outputs $y$ with $My+h=v^\star$.
By Theorem~\ref{thm:slice-layer-properties}, $\Psoc(f)$ is one such output, so the inner solution set is the nonempty closed affine set $Y^\star=\Psoc(f)+\ker M$.
The outer problem is the Euclidean projection of $f$ onto $Y^\star$, which is unique and is characterized by $z\in Y^\star$ and $z-f\in(\ker M)^\perp=\operatorname{range}M^\top$.
Since $\Psoc(f)-f=M^+(v^\star-w)\in\operatorname{range}M^+=\operatorname{range}M^\top$, the point $z=\Psoc(f)$ satisfies both conditions.
\hfill$\square$

\section{Proof of Theorem~\ref{thm:differentiability}}
\label{ssec:implicit-differentiation}

For fixed $x$, let $\mathcal P_{\mathrm{SOC},x}(y)$ denote the correction obtained from \eqref{eq:socnet-layer} by replacing $f_\theta(x)$ with $y$.
\begin{lemma}[Fixed-input stability]
    \label{lem:slice-layer-lipschitz}
    Fix $x$ with $\operatorname{rank}M(x)=r\ge1$.
    Then, for every $y_1,y_2\in\R^{\nout}$,
    \begin{equation}
        \label{eq:slice-layer-lipschitz}
        \|\mathcal P_{\mathrm{SOC},x}(y_1)-\mathcal P_{\mathrm{SOC},x}(y_2)\|_2\le\kappa_+(M(x))\|y_1-y_2\|_2,
    \end{equation}
    where $\kappa_+$ is defined in \eqref{eq:positive-condition-number}.
\end{lemma}

\noindent\emph{Proof.}
Let $\delta=y_1-y_2$ and define $w_i=M(x)y_i+h(x)$ and $v_i=\Pi_{\mathcal A(x)\cap\mathcal K}(w_i)$.
The difference of the corrected outputs is
\begin{equation}
    \label{eq:slice-layer-difference}
    \mathcal P_{\mathrm{SOC},x}(y_1)-\mathcal P_{\mathrm{SOC},x}(y_2)=\bigl(I-M(x)^+M(x)\bigr)\delta+M(x)^+(v_1-v_2).
\end{equation}
The two terms in \eqref{eq:slice-layer-difference} are orthogonal because they belong to $\ker M(x)$ and $\operatorname{range}M(x)^\top$, respectively.
Nonexpansiveness of the metric projector gives
\begin{equation*}
    \|M(x)^+(v_1-v_2)\|_2\le\frac{\|w_1-w_2\|_2}{\sigma_r(M(x))}=\frac{\|M(x)M(x)^+M(x)\delta\|_2}{\sigma_r(M(x))}\le\kappa_+(M(x))\,\|M(x)^+M(x)\delta\|_2.
\end{equation*}
Consequently,
\begin{equation*}
    \begin{aligned}
        \|\mathcal P_{\mathrm{SOC},x}(y_1)-\mathcal P_{\mathrm{SOC},x}(y_2)\|_2^2
        &\le\bigl\|\bigl(I-M(x)^+M(x)\bigr)\delta\bigr\|_2^2+\kappa_+(M(x))^2\|M(x)^+M(x)\delta\|_2^2\\
        &\le\kappa_+(M(x))^2\|\delta\|_2^2,
    \end{aligned}
\end{equation*}
since $\kappa_+(M(x))\ge1$.
Taking square roots proves the result.
\hfill$\square$

Lemma~\ref{lem:slice-layer-lipschitz} gives the Lipschitz bound in Theorem~\ref{thm:differentiability}, and Rademacher's theorem gives differentiability almost everywhere.
Differentiating \eqref{eq:socnet-layer} with respect to $f_\theta(x)$ by the chain rule gives \eqref{eq:slice-layer-jacobian}.

Let $G(\lambda,\vartheta)=0$ denote the active scalar equation in Table~\ref{tab:slice-roots}, i.e., $F_E$, $F_P$, or $H$, where $\vartheta$ collects the differentiable inputs to the slice geometry.
At every $w\notin\mathcal K$ with a nondegenerate slice, \eqref{eq:ellipsoid-monotonicity}, \eqref{eq:paraboloid-monotonicity}, and \eqref{eq:hyperboloid-monotonicity} give $\partial_\lambda G(\lambda^\star,\vartheta)<0$.
The implicit function theorem then gives
\begin{equation}
    \label{eq:implicit-root-derivative}
    D\lambda^\star[\dot\vartheta]=-\frac{D_\vartheta G(\lambda^\star,\vartheta)[\dot\vartheta]}{\partial_\lambda G(\lambda^\star,\vartheta)}.
\end{equation}
The derivative of the lifted projection follows by differentiating \eqref{eq:slice-projection} with the reconstruction in Table~\ref{tab:slice-roots}.
Thus, reverse-mode differentiation can differentiate the defining scalar equation without unrolling the numerical root solver.
\hfill$\square$

Classical differentiability can fail at the feasible boundary, the cone apex, a degenerate slice, or a change of slice type.
Differentiation with respect to $M(x)$ and $h(x)$ additionally requires locally constant rank and nonzero singular values bounded away from zero.

\section{Proof of Theorem~\ref{thm:uat}}
\label{app:uat-proof}

Constant rank and continuity of $M$ imply continuity of $M^+$ on $\mathcal X$.
Together with the continuity of the moving projector, this shows that $\Psoc(f)$ is continuous whenever $f$ is continuous.
Theorem~\ref{thm:slice-layer-properties} gives feasibility and identity on feasible functions.
The positive singular values of a continuous constant-rank matrix field vary continuously.
Compactness of $\mathcal X$ therefore implies that $\sigma_r(M(x))$ is bounded away from zero and that $\overline\kappa$ is finite.

Let $g$ be continuous and feasible.
Applying Lemma~\ref{lem:slice-layer-lipschitz} pointwise and using $\Psoc(g)=g$ gives
\begin{equation*}
    \|\Psoc(f)-g\|_\infty=\|\Psoc(f)-\Psoc(g)\|_\infty\le\overline\kappa\|f-g\|_\infty,
\end{equation*}
which is \eqref{eq:uat-bound}.
Since $\mathcal F$ is dense, for any $\epsilon>0$ there exists $f\in\mathcal F$ with $\|f-g\|_\infty<\epsilon/\overline\kappa$, and then $\|\Psoc(f)-g\|_\infty<\epsilon$, which proves the density claim.
\hfill$\square$

The continuity condition on the moving projector isolates the set-valued regularity required when the affine slice changes with $x$.
It can be verified from the explicit branch formulas under suitable nondegeneracy or from a separate continuity condition on the set map $x\mapsto\mathcal A(x)\cap\mathcal K$.

\section{Details of the contact force allocation experiment}
\label{app:friction}

\begin{figure}[!htbp]
\centering
\includegraphics[width=0.50\linewidth]{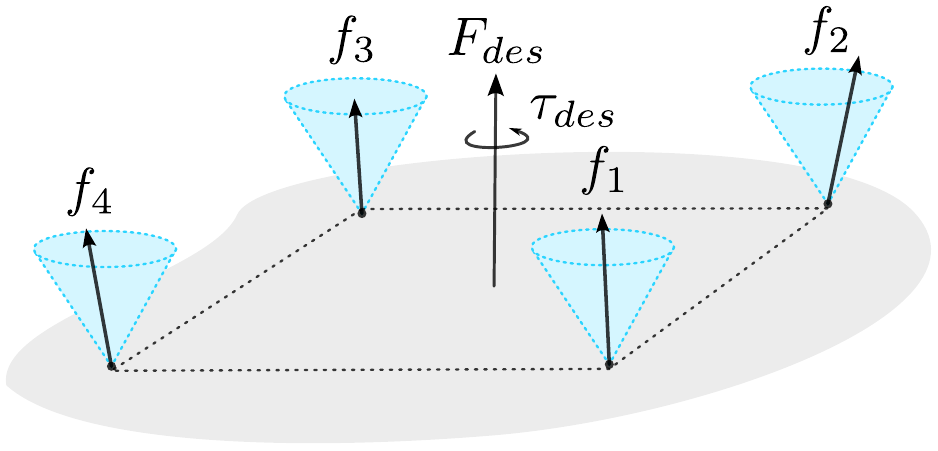}
\caption{Contact geometry of the force allocation task.}
\label{fig:friction}
\end{figure}

\paragraph{Problem data.}
The four contacts sit at the corners of a unit square centered at the object center, with normals along $+z$ (Fig.~\ref{fig:friction}).
The contact positions, friction coefficient, grasp map, and SOC lifting are
\begin{equation}
\begin{gathered}
r_1=\begin{bmatrix}0.5\\0.5\\0\end{bmatrix},\quad
r_2=\begin{bmatrix}0.5\\-0.5\\0\end{bmatrix},\quad
r_3=\begin{bmatrix}-0.5\\0.5\\0\end{bmatrix},\quad
r_4=\begin{bmatrix}-0.5\\-0.5\\0\end{bmatrix},\qquad
\mu_{\mathrm f}=0.6,\\[2pt]
Gy=\begin{bmatrix}\sum_{i=1}^4 f_i\\[3pt]\sum_{i=1}^4 r_i\times f_i\end{bmatrix},\qquad G\in\R^{6\times12},\\[2pt]
M_i=\begin{bmatrix}0_{3\times3(i-1)}&\bar M&0_{3\times3(4-i)}\end{bmatrix},\quad
\bar M=\begin{bmatrix}0&0&\mu_{\mathrm f}\\1&0&0\\0&1&0\end{bmatrix},\\[2pt]
M=[M_1;\ldots;M_4]=\operatorname{blkdiag}(\bar M,\ldots,\bar M),\quad
h=0,\quad
\mathcal K=\mathcal K_1\times\cdots\times\mathcal K_4.
\end{gathered}
\label{eq:contact-data}
\end{equation}

\paragraph{Data.}
Each target wrench is $x=Gy_{\mathrm{gen}}$ for a random allocation $y_{\mathrm{gen}}$ strictly inside the friction cones, so every target can be matched with zero loss.
Each contact force of $y_{\mathrm{gen}}$ is drawn independently, with $f_{iz}\sim\mathrm{Unif}(0.5,2)$ and a tangential component of uniformly random direction and magnitude $0.8\,\mu_{\mathrm f}f_{iz}\,\mathrm{Unif}(0,1)$.
The training, validation, and test sets contain $16{,}384$, $2{,}048$, and $4{,}096$ wrenches, and $y_{\mathrm{gen}}$ is never used as a training target.

\paragraph{Network and training.}
Every learned method uses the backbone MLP $6\to64\to64\to12$ with $\tanh$ activations and PyTorch's default initialization, initialized identically across methods for each seed.
Training minimizes the wrench MSE with Adam (learning rate $2\times10^{-3}$, batch size $128$) for $3{,}000$ updates, and the five seeds set the initialization and the batch order.
The validation wrench MSE is evaluated every $25$ updates, and the best checkpoint is kept.

\paragraph{Baselines.}
The penalty network adds $\lambda\sum_{i=1}^4\bigl[\|(f_{ix},f_{iy})\|_2-\mu_{\mathrm f}f_{iz}\bigr]_+^2$, averaged over the batch, to the loss, with $\lambda\in\{0.1,1,10,10^2\}$ selected by the lowest validation violation rate ($\lambda=10^2$).
The optimization references solve \eqref{eq:friction-task-main} for one input at a time through CVXPY with ECOS and Clarabel at tolerance $10^{-8}$.

\paragraph{Evaluation.}
The friction violation of an output is $\max_i\bigl[\|(f_{ix},f_{iy})\|_2-\mu_{\mathrm f}f_{iz}\bigr]_+$; SOC-Net's nonzero values are floating-point round-off from the pull-back by $M^{-1}$.
For the learned methods, $T_{\mathrm{test}}$ is the median over $200$ calls, after $20$ warm-up calls, of one batched forward pass on the whole test set; for the solvers, it is the median of three passes over the test set.
$T_{\mathrm{train}}$ is the wall time of the $3{,}000$ updates, excluding validation.

\section{Details of the 3-D obstacle-avoidance experiment}
\label{app:di3d}

\paragraph{Problem data.}
The obstacle is a sphere centered at $o=(-2.25,0.2,1.5)$ with radius $0.6\,\mathrm m$, which we inflate by the vehicle radius $0.15\,\mathrm m$ to $r_{\mathrm o}=0.75\,\mathrm m$; the goal is $p^\star=(0,0,1.5)$, $g=9.81\,\mathrm{m/s^2}$, and the disturbance bound is $\varepsilon_d=0.5$.
The cost \eqref{eq:di3d-cost} uses $Q_p=10I$, $Q_v=I$, $R=0.1I$, $N=200$, and $t_{\mathrm s}=0.02\,\mathrm s$; the input is held constant over each step, and the dynamics are integrated by RK4 with $10$ substeps.
The PD controller is the gravity-compensated law $u_{\mathrm{PD}}=ge_z-4(p-p^\star)-3v$, clipped to norm $2g$; without a safety filter, it collides from every test state.

\paragraph{Robust CBF constraint.}
The safe set is $\{\phi\ge0\}$ with $\phi(p)=\|p-o\|_2^2/r_{\mathrm o}^2-1$.
Because $\phi$ has relative degree two, the barrier is $B=\dot\phi+k_h\phi$, with $(k_h,k_B)=(3,8)$ chosen from $\{1,2,3\}\times\{3,5,8\}$ as the pair with the lowest validation cost among those for which PD\,+\,SOCP filter does not collide.
Along the dynamics, $\dot B=\ell(x)+c(x)^\top\bigl((I+\Delta)u+\delta\bigr)$ with
\begin{equation*}
c(x)=\frac{2(p-o)}{r_{\mathrm o}^2},
\qquad
\ell(x)=\frac{2\|v\|_2^2}{r_{\mathrm o}^2}+k_h\,c(x)^\top v-g\,c_z(x).
\end{equation*}
At a fixed state and commanded input,
\begin{equation*}
\inf_{\|\Delta\|_2\le\varepsilon_g}c^\top\Delta u=-\varepsilon_g\|c\|_2\|u\|_2,
\qquad
\inf_{\|\delta\|_2\le\varepsilon_d}c^\top\delta=-\varepsilon_d\|c\|_2,
\end{equation*}
attained at $\Delta=-\varepsilon_g cu^\top/(\|c\|_2\|u\|_2)$ and $\delta=-\varepsilon_d c/\|c\|_2$, so \eqref{eq:di3d-robust-soc} with $d(x)=\ell(x)+k_BB(x)-\varepsilon_d\|c(x)\|_2$ is the exact robust counterpart of $\dot B+k_BB\ge0$ for all admissible $(\Delta,\delta)$, i.e., the input-robust CBF condition of \citet{taylor2021towards}.
SOC-Net acts on the residual $y=f_\theta(x)$, with $u=u_{\mathrm{PD}}(x)+y$, through
\begin{equation*}
M(x)=\begin{bmatrix}c(x)^\top\\\varepsilon_g\|c(x)\|_2I_3\end{bmatrix}\in\R^{4\times3},
\qquad
h(x)=\begin{bmatrix}d(x)+c(x)^\top u_{\mathrm{PD}}(x)\\\varepsilon_g\|c(x)\|_2u_{\mathrm{PD}}(x)\end{bmatrix}.
\end{equation*}
For every $c(x)\ne0$, the slice has $s^2=1/(1+\varepsilon_g^2)$, $\mu=(\varepsilon_g^2-1)/(\varepsilon_g^2+1)$, and $\kappa_+=\sqrt{1+1/\varepsilon_g^2}$, none of which depends on the state.
We write $r_{\mathrm{CBF}}(x,u)=c(x)^\top u+d(x)-\varepsilon_g\|c(x)\|_2\|u\|_2$ for the signed robust margin.

\paragraph{Feasibility and safety.}
On the safe set, $\|p-o\|_2\ge r_{\mathrm o}$, so $\|c(x)\|_2\ge2/r_{\mathrm o}>0$ and $M(x)$ has full column rank.
Since $\varepsilon_g<1$, the input $u=\beta c(x)$ gives $r_{\mathrm{CBF}}=\beta(1-\varepsilon_g)\|c(x)\|_2^2+d(x)>0$ for large enough $\beta$, so Assumption~\ref{ass:feasibility} holds at every safe state.
Every start has $v_0=0$ and $\phi(p_0)>0$, so $B(x_0)=k_h\phi(p_0)>0$; because SOC-Net satisfies \eqref{eq:di3d-robust-soc} at every evaluation (Theorem~\ref{thm:slice-layer-properties}), the standard higher-order CBF argument gives forward invariance of $\{\phi\ge0\}$ when the constraint is enforced in continuous time.
The experiments enforce it only at the control instants under a zero-order hold; still, over all visited states $\|c(x)\|_2\ge4.05>2/r_{\mathrm o}=2.67$, every active query took the hyperboloid branch, no constrained method collided, and every learned constrained policy kept at least $0.36\,\mathrm m$ of clearance.

\paragraph{Data.}
Initial positions are drawn uniformly from the box $[-5,-4.5]\times[-0.5,0.5]\times[1,2]$ with $v_0=0$ and rejection-sampled for $\phi(p_0)>0$; the obstacle blocks every straight path from this box to the goal.
The training, validation, and test sets contain $800$, $100$, and $100$ initial states.

\paragraph{Network and training.}
Every learned method uses the backbone MLP $9\to200\to200\to3$ with ReLU activations on the input $e(x)=(p-p^\star,v,p-o)$, initialized identically across methods for each seed.
Training rolls out the nominal dynamics ($\Delta=0$, $\delta=0$) and backpropagates $J$ through the rollout and the constraint layer, using Adam (learning rate $3\times10^{-3}$, batch size $100$) for $200$ epochs ($1{,}600$ updates); the five seeds set the initialization and the batch order.
The validation cost is evaluated every $10$ epochs, and the best checkpoint is kept.
The learning rate was chosen from $\{10^{-4},3\times10^{-4},10^{-3},3\times10^{-3}\}$ on the unconstrained network's validation cost.

\paragraph{Baselines.}
The penalty network adds $\lambda\sum_k t_{\mathrm s}\,[-r_{\mathrm{CBF}}(x_k,u_k)]_+^2$ to $J$, with $\lambda\in\{1,10,10^2,10^3\}$ selected by the lowest validation violation rate ($\lambda=10^3$).
RAYEN uses the interior anchor $\beta(x)c(x)$ with $\beta(x)=\max\{0,-d(x)/((1-\varepsilon_g)\|c(x)\|_2^2)\}+0.1$, $\Pi$Net runs $100$ iterations with an unrolled backward pass, and HardNet-Cvx computes \eqref{eq:projection-target} with cvxpylayers.
LMI-Net, which has no public code, is our implementation with $500$ Douglas--Rachford iterations.
The SOCP filters solve $\min_u\frac12\|u-u_{\mathrm{ref}}\|_2^2$ subject to \eqref{eq:di3d-robust-soc} through CVXPY with Clarabel at tolerance $10^{-10}$, where $u_{\mathrm{ref}}$ is the output of the trained unconstrained network or the PD controller $u_{\mathrm{PD}}$.

\paragraph{Evaluation.}
The $102$ scenarios are the nominal one, a state-dependent worst case, and $100$ random ones shared by all methods.
The worst case applies $\tilde u=u-(\varepsilon_g\|u\|_2+\varepsilon_d)c/\|c\|_2$, recomputed at every substep; each random scenario draws $\Delta=\varepsilon_g\rho\,G/\|G\|_2$, with $G$ a $3\times3$ standard normal matrix, and $\delta=\varepsilon_d\rho' n$, with $n$ uniform on the unit sphere and $\rho,\rho'\sim\mathrm{Unif}(0,1)$, and keeps them fixed over the rollout.
A rollout violates the constraint if $r_{\mathrm{CBF}}<-10^{-9}\max\{|d|,\|c\|_2\|u\|_2,1\}$ at some control instant, and its CBF violation is the largest $[-r_{\mathrm{CBF}}]_+$ over its control instants; collisions are checked on a grid ten times finer than the substeps.
$T_{\mathrm{test}}$ is the median wall time over $200$ calls, after $20$ warm-up calls, of one inference of the full policy (backbone and constraint layer or filter) on a batch of $1024$ states sampled from SOC-Net's nominal test rollouts; the solver calls of HardNet-Cvx and the SOCP filters run on the CPU inside this timed call.
SOC-Net is compiled with \texttt{torch.compile}, and the other layers except LMI-Net run as released; HardNet-Cvx and the filters call the solver only when the raw output is infeasible, since a projection, like SOC-Net, leaves feasible outputs unchanged.
$T_{\mathrm{train}}$ is $200$ times the median wall time of one training epoch, excluding validation.

\end{document}